\documentclass{article}
\usepackage{spconf,amsmath,graphicx}
\usepackage{booktabs}
\usepackage{enumitem}
\setlist{nosep, leftmargin=12pt}
\usepackage{xcolor}
\definecolor{swiftblue}{rgb}{0.21,0.49,0.74}
\usepackage[breaklinks,colorlinks,allcolors=swiftblue]{hyperref}
\hypersetup{pdftitle={SWIFT-Cell: Segmentation Without Iterative Flow Tracking},
  pdfauthor={Devansh Lalwani, Swapnil Bhat, Muralikrishna H H, Anand Reddy, Maulik Shah}}
\usepackage[capitalize,nameinlink]{cleveref}
\crefname{section}{Sec.}{Secs.}
\Crefname{section}{Sec.}{Secs.}


\title{SWIFT-Cell: Segmentation Without Iterative Flow Tracking}

\name{Devansh Lalwani$^{1}$, Swapnil Bhat$^{1}$, Muralikrishna H H$^{2}$, Anand Reddy$^{1}$, Maulik Shah$^{1}$}
\address{\small $^{1}$Turocrates.ai, Mumbai, India\quad $^{2}$Department of Pathology, ESIC Medical College and Hospital, Kalaburagi, Karnataka, India}

\begin{document}
\maketitle

\begin{abstract}
Generalist segmenters such as Cellpose-SAM (CPSAM) outline cells accurately across stains but slowly; watershed detectors such as QuPath's are fast but less accurate.
In time-critical use, such as interactive retrieval of cells from a user-drawn region, CPSAM at the reference settings takes 58--303\,s per measured $8192^2$ region on an RTX 4500 Ada GPU, beyond a seconds-scale interactive target.
We distil CPSAM into SWIFT-Cell (SWIFT), a 0.63M-parameter U-Net trained only on CPSAM masks of 1,202 H\&E and IHC tiles.
On three $8192^2$ development regions SWIFT takes 2.0--5.2\,s on 4 CPU cores in int8, 2.5--5.1$\times$ faster than QuPath's detector on the same cores, and 0.36--0.49\,s on a GPU.
Against human annotations on eight H\&E, IHC and ISH sets it keeps 95\% of CPSAM's panoptic quality (PQ) on PanNuke (0.593 vs.\ 0.627) and is level on NuCLS (0.546 vs.\ 0.540). On Ki-67 IHC masks scored to the outline centre it is 0.205 PQ above QuPath's better mode, 0.079 below CPSAM and 0.027 below InstanSeg (CI includes 0; at the outer edge 0.008 below CPSAM); on 212 pathologist-outlined nuclei in Ki-67 fields of 12 untouched patients its recall is 0.925 (CPSAM 0.920, InstanSeg 0.901; paired difference CIs include 0); one method-blinded pathologist accepted its outlines in 96\% of 100 fields (60 from two development tissue patients, 40 hematology; CPSAM 83\%, QuPath 53\%; tissue 100\% for SWIFT and CPSAM); on ER/PR points from 12 patients its detection F1 is 0.882 (CPSAM 0.896, InstanSeg 0.857), and on 4-reader PD-L1 points 0.83 (CPSAM 0.85, readers 0.87).
SWIFT is close to CPSAM in accuracy and faster than watershed detection: it bridges the gap between research and real-world deployment.
\end{abstract}


\section{Introduction}
\label{sec:intro}
Generalist segmenters such as Cellpose-SAM (CPSAM)~\cite{pachitariu2025cellposesam} work across stains and modalities without retraining.
Some uses are time-critical: a user draws a region and waits for its cells to be outlined, a deployment has only CPU cores or a T4-class GPU, or whole regions must be segmented within seconds.
On the timed regions (\cref{tab:main}), CPSAM on a GPU at QuPath's working resolution of 0.5\,\textmu m/px takes 3.3--4.9$\times$ as long as QuPath on 4 CPU cores.

QuPath's cell detector~\cite{bankhead2017qupath} takes seconds on a CPU. It thresholds one colour-deconvolved channel and splits nuclei by watershed. Because detection runs on the hematoxylin channel by default, DAB-positive nuclei are often missed or outlined smaller than by CPSAM (\cref{sec:results}).
Lighter supervised nucleus models exist (StarDist~\cite{schmidt2018stardist}, InstanSeg~\cite{goldsborough2024instanseg}), and students have been trained on a teacher's output~\cite{hinton2015distilling}: HoverFast~\cite{liakopoulos2024hoverfast} (a 0.94M-parameter U-Net) and HoVer-UNet~\cite{tommasino2024hoverunet} learn from HoVer-Net~\cite{graham2019hovernet} outputs, \cite{shvetsov2025lightweight} distils a pathology foundation model, and EfficientBioAI~\cite{zhou2024efficientbioai} quantises Cellpose to int8. We found no prior distillation of CPSAM.

Large generalists are the research state of the art, but real-world deployment has time and hardware limits. To bridge this gap, we distil the trusted generalist on its own masks into a student sized for those limits, and measure the cost.

\noindent\textbf{Contributions.}
\begin{enumerate}
\item \textbf{A recipe.} Distil the trusted generalist into a 0.63M-parameter U-Net using only its masks on the target stain mix, decode in one pass from centre offsets~\cite{cheng2020panoptic}, and run it in fp16 on a GPU or int8 on a CPU (\cref{sec:method}).
\item \textbf{A measured trade-off.} Speed, fidelity to the teacher and DAB-positive outline size against QuPath, InstanSeg and the teacher on the same hardware (\cref{tab:main}).
\item \textbf{An evaluation protocol.} What the student keeps and loses against the teacher, human annotations on H\&E and IHC (InstanSeg, StarDist and QuPath as references), zero-shot hematology and a blinded pathologist rating (\cref{tab:gt}, \cref{sec:heme}).
\end{enumerate}

\section{Distilling the teacher}
\label{sec:method}
\noindent\textbf{Pseudo-labels.}
We sampled 1,202 tiles of about $307{\times}307$\,\textmu m ($1024^2$ px at 0.30\,\textmu m/px): 916 from 249 institutional whole-slide images (H\&E of several organs; Ki-67, CD3/CD8, ER, PR, HER2 and other IHC), 168 from 43 TCGA H\&E slides~\cite{weinstein2013tcga} and 118 from 30 ACROBAT slides~\cite{weitz2024acrobat} (ER, PR, HER2).
CPSAM labelled every tile with the reference settings (\cref{sec:results}). No human labels were used to train the student.
No IGNITE~\cite{spronck2025ignite} data was used in training, warm start or int8 calibration.

\noindent\textbf{Student.}
The student is a 5-level U-Net~\cite{ronneberger2015unet} with 16 base channels, additive skips and a dilated bottleneck: 0.63M parameters (0.6\,MB in int8), 487$\times$ fewer than CPSAM (304.6M).
Input is resampled to 0.5\,\textmu m/px; a stride-2 stem places the finest feature grid at 1\,\textmu m/px.
A full-resolution stage (adding 928 parameters) refines the decoder features on the input grid, where three outputs are predicted: a foreground logit (BCE + Dice), a Gaussian centre heatmap ($\sigma{=}1.5$\,px, weighted MSE) and the offset $(\Delta y,\Delta x)$ from each foreground pixel to its centroid (smooth-L1, weight 2).
Training uses AdamW, flip, rotation, colour, blur and scale augmentation, and up-weights the palest third of hematoxylin nuclei. A first model is trained in two stages on an earlier set of up to 811 tiles (12k steps, 70\,min on one GPU), then fine-tuned on the full set (1,091 training, 111 validation tiles; 10k steps, 40\,min); the last checkpoint is used. Some validation tiles may have been warm-start training tiles, so validation did not select the model and is not independent. The full-resolution stage, stem and pale weighting were not ablated.

\noindent\textbf{Offset decoding.}
Seeds are local maxima of the heatmap ($7{\times}7$ max-pool, threshold 0.35) inside the foreground (probability $>0.5$) on the 0.5\,\textmu m/px grid; the seed map $V$ is dilated by 5\,px (GPU) or 3\,px (CPU), and where dilations overlap the larger ID wins.
Foreground and offsets are bilinearly resampled to the output grid. An output pixel $p$ with working-grid position $\pi(p)$ is foreground if $f(\pi(p))>0.5$ and takes the label $V\big(\mathrm{round}(\pi(p)+o(\pi(p)))\big)$, so a pixel that reaches no seed stays background. Objects under 5 working-grid pixels are removed.
Outputs are computed per tile with a 48-px halo; the halo-free centres form one region map, decoded once (no per-tile instance reconciliation), and empty glass tiles are skipped. The GPU path runs in fp16; the CPU path uses OpenVINO int8, post-training quantised on 64 training tiles, in throughput mode with asynchronous requests.

\section{Results on tissue}
\label{sec:results}
\noindent\textbf{Setup.}
We use three large regions: H\&E breast (27k CPSAM detections), Ki-67 (97k) and CD3 colon (10k). They lie on development slides: their crops tuned SWIFT's decoding, and the regions compared design variants. A warm-start set filtered by name held 4 tiles from a renamed copy of the Ki-67 slide, outside its region.
CPSAM uses these reference settings: 0.30\,\textmu m/px, $1536^2$ blocks, cellprob $-1$, flow threshold 0.6. QuPath 0.7.0 runs headless with defaults (\texttt{WatershedCellDetection} at 0.5\,\textmu m/px, default stain vectors, measurements and 5\,\textmu m cell expansion).
CPU work is pinned to 4 physical cores of an AMD EPYC 7282; GPU runs use an RTX 4500 Ada plus the same cores. Segmentation latency is measured in one session on a decoded RGB region, without slide reading or model loading: median of 5 runs after a warm-up (CPSAM, InstanSeg CPU: 3). For SWIFT and InstanSeg it includes resampling, host-device copies and upsampling of labels to the native grid, for QuPath measurements and cell expansion; for CPSAM it excludes painting masks into a native-grid label image.
Fidelity is object F1 at IoU${\geq}0.5$ against reference CPSAM (qualifying pairs verified one-to-one); it measures agreement with the teacher, not accuracy.

\begin{table}[t]
\centering\small
\setlength{\tabcolsep}{2.5pt}
\resizebox{\columnwidth}{!}{%
\begin{tabular}{@{}lccc@{}}
\toprule
Method (hardware) & Time (s) & F1 vs.\ CPSAM & DAB$^+$\\
 & H\&E / Ki-67 / CD3 & H\&E / Ki-67 / CD3 & area\\
\midrule
CPSAM 0.30\,\textmu m/px (GPU) & 86 / 303 / 58 & 1 / 1 / 1 & 1.00\\
CPSAM 0.5\,\textmu m/px (GPU) & 34 / 131 / 23 & .93 / .95 / .84 & 1.00\\
QuPath (4 CPU thr.) & 10.2 / 26.6 / 5.1 & .68 / .54 / .61 & 0.68\\
InstanSeg~\cite{goldsborough2024instanseg} (GPU fp16) & 1.73 / 5.95 / 1.19 & .87 / .84 / .76 & 0.85\\
InstanSeg (4 CPU thr.) & 78 / 199 / 49 & .88 / .84 / .76 & 0.85\\
\midrule
\textbf{SWIFT} (GPU fp16) & 0.38 / 0.49 / 0.36 & .89 / .88 / .73 & 1.00\\
\textbf{SWIFT} (4 CPU int8) & 2.32 / 5.23 / 2.02 & .89 / .87 / .73 & 0.97\\
\bottomrule
\end{tabular}}
\caption{Evaluation regions on development slides. Times: one session (\cref{sec:results}). DAB$^+$ area: median area ratio to CPSAM, matched DAB-positive Ki-67 nuclei. InstanSeg: released \texttt{brightfield\_nuclei} v0.1.1 at 0.5\,\textmu m/px, antialiased, our tiling; GPU: its TorchScript decoder, CPU: ours. SWIFT GPU (fp16, seed dilation 5\,px) and CPU (int8, 3\,px) are two deployment configurations, not a quantisation comparison.}
\label{tab:main}
\end{table}

\begin{table*}[t]
\centering\scriptsize
\setlength{\tabcolsep}{2.4pt}
\begin{tabular}{@{}llcrlcccccl@{}}
\toprule
Set & Stain, tissue; reference & Patients & Nuclei & Metric & SWIFT & CPSAM & InstanSeg & StarDist & QuPath & $\Delta$ vs.\ CPSAM [95\% CI]\\
\midrule
PanNuke fold 1~\cite{gamper2019pannuke} & H\&E, 19 tissues; masks & no ids & 63,218 & PQ & .593 & .627 & .589 & .577 & .417 & $-$.033 [$-$.036, $-$.030]\\
NuCLS~\cite{amgad2022nucls} & H\&E, breast; polygons & 122 & 19,175 & PQ & .546 & .540 & .524 & .527 & .413 & $+$.006 [.000, .011]\\
MoNuSeg test~\cite{kumar2017dataset} & H\&E, 7 organs; masks & 14 & 6,697 & PQ & .629 & .656 & .657 & .665 & .503 & $-$.027 [$-$.039, $-$.010]\\
BC-DeepLIIF~\cite{ghahremani2022deepliif} & Ki-67, breast; tumour masks & unknown (10--15 inferred) & 6,240 & PQ & .609 & .688 & .637 & .496 & .344 (.405) & $-$.079 [$-$.089, $-$.064]\\
EndoNuke~\cite{naumov2022endonuke} & ER/PR, endometrium; points & 12 & 210,419 & F1 & .882 & .896 & .857 & .684 & .803 & $-$.014 [$-$.019, $-$.005]\\
IGNITE, 4 readers~\cite{spronck2025ignite} & PD-L1, lung; points & 10 & 9,676 & F1 & .826 & .851 & n.m. & n.m. & .657 & $-$.025 [$-$.062, $-$.005]\\
Ki67-IH12 (in-house) & Ki-67, breast; outlined nuclei & 12 & 212 & Recall & .925 & .920 & .901 & .703 & .882 (.896) & $+$.005 [$-$.020, $+$.040]\\
\bottomrule
\end{tabular}
\caption{Human ground truth across stains, with patient counts where known. Pooled binary PQ~\cite{kirillov2019panoptic} (IoU${>}0.5$) for masks, detection F1 or recall (${\leq}4$\,\textmu m, 1:1) for points; no tuning on these sets. SWIFT: GPU fp16 (int8 within 0.004). QuPath in parentheses: OD-sum mode (threshold 0.1, fixed on development data). BC-DeepLIIF: 41 fields; predictions ${\geq}50$\% outside all masks dropped. EndoNuke: 1,472 of 1,740 tiles by non-pathologists. IGNITE: 32 test ROIs, reader mean (inter-reader F1 0.867). Ki67-IH12: recall of 212 pathologist-outlined nuclei (centroids). InstanSeg: \cref{tab:main} settings (CPU on PanNuke and NuCLS; released pipeline within 0.007 except NuCLS, .475). CIs: paired bootstrap over patients, PanNuke patches or BC-DeepLIIF inferred cases. n.m.: not measured.}
\label{tab:gt}
\end{table*}

\begin{figure}[t]
\centering
\includegraphics[width=0.85\columnwidth]{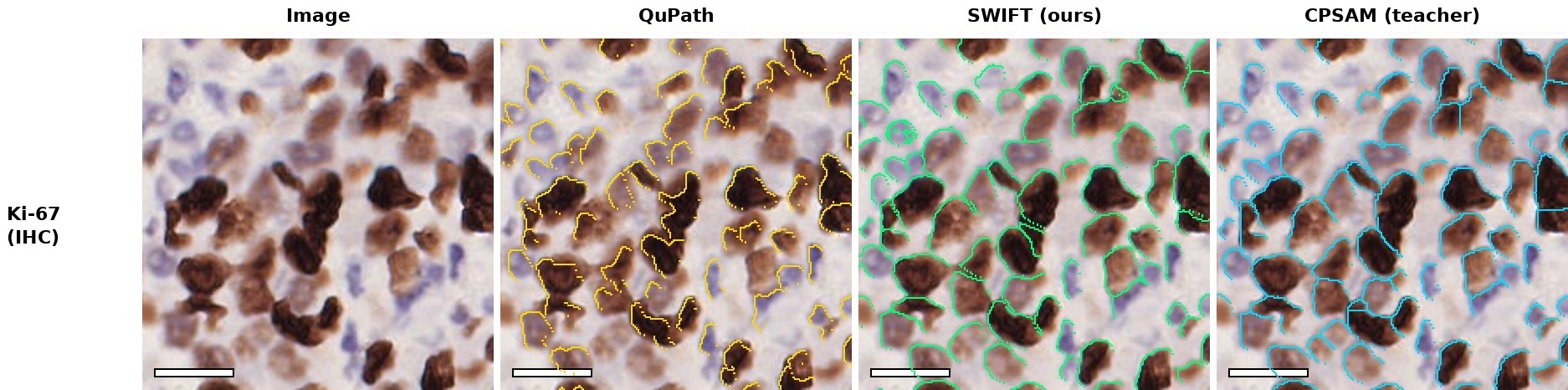}
\caption{Ki-67 tuning crop (development slide). Scale bar 20\,\textmu m.}
\label{fig:qual}
\end{figure}

\noindent\textbf{What is kept and what is lost.}
\begin{itemize}
\item \emph{Speed:} against QuPath detection alone (F1 0.94--0.97 to its default output) the int8 segmenter is 1.7--3.0$\times$ faster on the same cores.
\item \emph{Ki-67:} of CPSAM's DAB-positive nuclei, default QuPath misses 39\% outright (no object at the CPSAM centroid) and outlines a further 20\% at under half their area; the student misses 2.2\% and recovers 91\% at IoU${\geq}0.5$, against 32\% for default QuPath and 76\% (area 0.74) for its optical-density-sum (OD-sum) mode (\cref{fig:qual}).
The labelling index does not gain: on this one region (CPSAM 45\%) the student's is 2.1 points lower, as 12\% of its objects are small, mostly DAB-negative extras; QuPath's is within 0.1 points, as misses and extras cancel.
\end{itemize}

\begin{table}[t]
\centering\footnotesize
\setlength{\tabcolsep}{1.8pt}
\begin{tabular}{@{}lccc@{}}
\toprule
Decoder & F1 vs.\ CPSAM & PQ & Flips (\%)\\
 & H\&E/Ki-67/CD3 & PanNuke/NuCLS & H\&E/Ki-67/CD3\\
\midrule
\textbf{Offsets, output} & .886/.879/.728 & .593/.546 & 6.0/25.5/7.2\\
Offsets, 0.5\,\textmu m & .886/.876/.727 & .587/.544 & 20.2/34.8/21.5\\
Heatmap WS, output & .882/.864/.726 & .590/.544 & 8.0/36.3/8.4\\
Nearest seed & .872/.842/.719 & .571/.535 & 18.5/53.3/17.8\\
\bottomrule
\end{tabular}
\caption{Decoder ablation: final model, GPU, same seeds and foreground (bold: SWIFT). Output / 0.5\,\textmu m: labels assigned on the native / working grid. WS: watershed. Flips: matched cells whose circularity-filter decision (${\geq}0.7$, shapes on one 0.30\,\textmu m/px grid) differs from CPSAM's.}
\label{tab:abl}
\end{table}

\noindent\textbf{Decoder ablation.}
Offsets help assignment and the output grid helps shape (\cref{tab:abl}): a heatmap watershed on the output grid comes within 0.004 of offsets except on dense Ki-67, and labels on the 0.5\,\textmu m/px grid keep F1 within 0.003 but flip 20--35\% of circularity decisions.

\noindent\textbf{Human ground truth.}
Patches (PanNuke: 2,656) are mirror-padded by 16\,\textmu m and packed into mosaics; an object is kept if at least 40\% of it lies inside its patch (\cref{tab:gt}). On NuCLS (1,720 fields) box and point annotations are ignored and predictions at least 50\% inside them are dropped; fields of two patients with possible training overlap are excluded. CPSAM's training includes 6,053 of 7,898 PanNuke tiles and MoNuSeg train, and StarDist's (\texttt{2D\_versatile\_he}, run at 0.25\,\textmu m/px) MoNuSeg train, so CPSAM's PanNuke cell is not an unseen-data score and both MoNuSeg test cells are in-domain; no compared model lists any other set of \cref{tab:gt}.
On NuCLS the student's F1 is 0.018 below CPSAM's [$-0.024$, $-0.012$]. StarDist has a higher NuCLS F1 than the student ($-0.027$ [$-0.040$, $-0.015$]).
On PD-L1 the student is lower than CPSAM for 8 of 10 patients and for each reader.
On MoNuSeg test (none of its 14 patients in SWIFT's training) InstanSeg and StarDist score 0.028 [0.018, 0.036] and 0.036 [0.026, 0.043] PQ above the student, with higher precision (0.87, 0.89 vs.\ 0.77) and lower recall (0.84 vs.\ 0.90).

\noindent\textbf{Untouched Ki-67 (Ki67-IH12).} Two $56{\times}56$\,\textmu m fields per patient were fixed before any model ran on Ki-67 slides of 12 breast carcinoma patients (distinct in the laboratory system; not in training, tuning or calibration) at the scanner's 0.44\,\textmu m/px. A pathologist blind to model output outlined 212 nuclei in them. Only these nuclei were compared: we report their recall, matching outline centroids one-to-one within 4\,\textmu m. SWIFT recalls 0.925 [0.877, 0.967] (CPSAM 0.920, InstanSeg 0.901, default QuPath 0.882, StarDist 0.703; paired difference CIs versus CPSAM and InstanSeg include 0) from 450 objects (CPSAM 420, QuPath 414, InstanSeg 358). It misses 14 of the 100 DAB-positive nuclei (mean DAB optical density above 0.3; CPSAM 13, InstanSeg 14, QuPath 16) and 2 of the 112 negative (CPSAM 4, InstanSeg 7, QuPath 9). Matched outlines are 1.5$\times$ the pathologist's area (CPSAM 1.6, InstanSeg 1.4, QuPath 1.1), so 110 of its 196 detected nuclei match at IoU${>}0.5$ (CPSAM 108, InstanSeg 132, QuPath 120).

\noindent\textbf{Independent IHC.}
BC-DeepLIIF~\cite{ghahremani2022deepliif} masks (annotator not stated) cover mainly tumour nuclei (BCData's consensus points mark tumour cells only); one instance per drawn interior, boundary at the outline centre; 0.35\,\textmu m/px was estimated from nucleus size before any model ran. CIs resample 15 inferred source cases.
SWIFT's PQ is 0.205 [0.155, 0.280] above QuPath OD-sum, with Ki-67$^+$/$^-$ recall 0.88/0.83 (CPSAM 0.91/0.84, OD-sum 0.80/0.40). Against InstanSeg its PQ is 0.027 lower ([$-0.057$, $+0.012$]) and its F1 0.054 higher ([0.018, 0.099]): InstanSeg's tighter outlines fit the outline-centre boundary (SQ 0.78 vs.\ 0.70), but it misses more Ki-67$^-$ nuclei (recall 0.65); scored to the outline's outer edge, SWIFT, InstanSeg and CPSAM reach PQ 0.698, 0.565 and 0.706.
Splits per 100 nuclei were 3.1/1.3/0.7/5.2 for SWIFT/CPSAM/InstanSeg/QuPath; merge rates were 4.0/4.0/2.4\% for the first three.
One method-blinded pathologist judged 58\% [49, 67] of SWIFT CPU int8's and 40\% [34, 45] of OD-sum's sampled excluded objects real nuclei (144 per method; 12 images, one per inferred case). Mean image-level boundary ratings were 4.5 vs.\ 2.4 of 5 (paired difference $+2.1$ [$+1.7$, $+2.5$]; 95\% CIs bootstrapped over images).
On EndoNuke ER/PR points InstanSeg's precision is higher than the student's (0.93 vs.\ 0.87) and its recall lower (0.79 vs.\ 0.89); on 40 tiles against 3 pathologists the student reaches 0.902 (CPSAM 0.921), and the pathologists agree at 0.914.
On SiNuS~\cite{hossain2026sinus} HER2 dual-ISH (21 patients; singular nuclei only; precision not measurable) SWIFT recalls 86\% of 1,292 consensus masks (int8 82\%), against CPSAM 99.8\%, InstanSeg 37\%, StarDist 25\% and QuPath 15\%, but on pale nuclei it can outline the ISH signal clusters, splitting 14--16\% of nuclei.

\section{Zero-shot hematology}
\label{sec:heme}
With no hematology training or tuning, we ran the student with its tissue settings as an exploratory test on automatically chosen fields of held-out peripheral smears and bone-marrow aspirate (BMA). On smears it splits 70\% of 84 CPSAM leukocytes into two or more objects (F1 against CPSAM's leukocyte objects 0.19, 95\% CI 0.13--0.27).
On 40 BMA counting fields of 2 held-out patients, the student covers 76\% of 1,985 compact chromatin blobs found without any segmenter (QuPath 58\%, the gain mostly in one patient; CPSAM 26\% at the reference settings, at most 78\% in a prespecified sweep and 84\% on grayscale input), a proxy that ignores false detections, merges and boundaries.
A hematology student trained the same way on CPSAM whole-cell masks (511 tiles, 31 patients) agrees with grayscale-input CPSAM at F1 0.73 on 5 held-out marrow patients (no human reference). The recipe needs no new labels but inherits the teacher's reliability: check the teacher first.

\noindent\textbf{Pathologist rating.} One method-blinded pathologist rated 100 fields of $200^2$\,\textmu m: 60 tissue fields from 2 development patients and 40 hematology fields in 7 clusters, each shown with SWIFT, reference-settings CPSAM and QuPath outlines. Acceptability for retrieval or counting was 96\% for SWIFT (field-level Wilson 95\% CI [90, 98]), 83\% for CPSAM and 53\% for QuPath; tissue rates were 100, 100 and 72\%, hematology rates 90, 58 and 25\%. SWIFT minus CPSAM was $+13$ points (field-level 95\% CI [$+6$, $+21$]; 9-cluster sensitivity CI [$+1$, $+41$]). Both intervals clear the prespecified $-10$-point margin, but few-cluster intervals may under-cover, and non-inferiority on a tissue population is not established. Mean quality (1--5) was 4.17, 3.95 and 2.78; the clustered SWIFT minus CPSAM CI was [$-0.03$, $+0.76$]. On 10 hidden repeats, excluded from efficacy, acceptability agreed 9 times and quality weighted $\kappa$ was 0.61. These results describe one reader's judgements on these fields.

\section{Limitations and conclusion}
\label{sec:limits}
The student depends on its teacher's masks: agreement with CPSAM does not establish accuracy where CPSAM is unreliable, as in a dense CD3 aggregate or, at the reference settings, marrow precursors. Its teacher PQ deficits are 0.033 on PanNuke, 0.027 on MoNuSeg and 0.079/0.008 on BC-DeepLIIF at the outline centre/outer edge (\cref{tab:gt}; on MoNuSeg InstanSeg and StarDist also score higher), and it changes CPSAM's circularity-filter decision for 6--26\% of matched cells (\cref{tab:abl}).
The timed regions lie on development slides. BC-DeepLIIF patients are inferred, its boundary convention is not stated and changes the InstanSeg ranking, and CD3 has no human-mask evidence. Times come from 4 EPYC cores and an RTX 4500 Ada, not a T4, and downstream retrieval performance is not measured.

Distilling a research generalist on its own masks can bridge the gap to real-world deployment under time and hardware limits: the 0.63M-parameter student runs faster than QuPath's watershed detector on the same CPU and stays close to the teacher in accuracy, with measured losses.

\noindent\textbf{Compliance with ethical standards.} Institutional slides were de-identified and used retrospectively under an ethics waiver; all other data are public datasets used under their licences.
\textbf{Conflict of interest.} The authors build solutions in the digital pathology space.

\bibliographystyle{IEEEbib}
{\footnotesize\setlength{\itemsep}{0pt}\bibliography{main}}

\begin{thebibliography}{10}

\bibitem{pachitariu2025cellposesam}
M.~Pachitariu et~al.,
\newblock ``Cellpose-{SAM}: superhuman generalization for cellular
  segmentation,''
\newblock {\em bioRxiv 2025.04.28.651001}, 2025.

\bibitem{bankhead2017qupath}
P.~Bankhead et~al.,
\newblock ``{QuPath}: Open source software for digital pathology image
  analysis,''
\newblock {\em Sci. Rep.}, vol. 7, 2017,
\newblock Art. no. 16878.

\bibitem{schmidt2018stardist}
U.~Schmidt et~al.,
\newblock ``Cell detection with star-convex polygons,''
\newblock in {\em MICCAI}, 2018.

\bibitem{goldsborough2024instanseg}
T.~Goldsborough et~al.,
\newblock ``{InstanSeg}: an embedding-based instance segmentation algorithm
  optimized for accurate, efficient and portable cell segmentation,''
\newblock {\em arXiv:2408.15954}, 2024.

\bibitem{hinton2015distilling}
G.~Hinton et~al.,
\newblock ``Distilling the knowledge in a neural network,''
\newblock {\em arXiv:1503.02531}, 2015.

\bibitem{liakopoulos2024hoverfast}
P.~Liakopoulos et~al.,
\newblock ``{HoverFast}: an accurate, high-throughput, clinically deployable
  nuclear segmentation tool for brightfield digital pathology images,''
\newblock {\em J. Open Source Softw.}, vol. 9, 2024,
\newblock Art. no. 7022.

\bibitem{tommasino2024hoverunet}
C.~Tommasino et~al.,
\newblock ``{HoVer-UNet}: Accelerating {HoVerNet} with {UNet}-based multi-class
  nuclei segmentation via knowledge distillation,''
\newblock in {\em ISBI}, 2024.

\bibitem{graham2019hovernet}
S.~Graham et~al.,
\newblock ``{HoVer-Net}: Simultaneous segmentation and classification of nuclei
  in multi-tissue histology images,''
\newblock {\em Med. Image Anal.}, vol. 58, 2019,
\newblock Art. no. 101563.

\bibitem{shvetsov2025lightweight}
N.~Shvetsov et~al.,
\newblock ``A lightweight and extensible cell segmentation and classification
  model for {H\&E}-stained cancer whole slide images,''
\newblock {\em Comput. Biol. Med.}, vol. 199, 2025,
\newblock Art. no. 111326.

\bibitem{zhou2024efficientbioai}
Y.~Zhou et~al.,
\newblock ``{EfficientBioAI}: making bioimaging {AI} models efficient in energy
  and latency,''
\newblock {\em Nat. Methods}, vol. 21, no. 3, pp. 368--369, 2024.

\bibitem{cheng2020panoptic}
B.~Cheng et~al.,
\newblock ``{Panoptic-DeepLab}: A simple, strong, and fast baseline for
  bottom-up panoptic segmentation,''
\newblock in {\em CVPR}, 2020.

\bibitem{weinstein2013tcga}
J.~N. Weinstein et~al.,
\newblock ``The {Cancer Genome Atlas} pan-cancer analysis project,''
\newblock {\em Nat. Genet.}, vol. 45, no. 10, pp. 1113--1120, 2013.

\bibitem{weitz2024acrobat}
P.~Weitz et~al.,
\newblock ``The {ACROBAT} 2022 challenge: Automatic registration of breast
  cancer tissue,''
\newblock {\em Med. Image Anal.}, vol. 97, 2024,
\newblock Art. no. 103257.

\bibitem{spronck2025ignite}
J.~Spronck et~al.,
\newblock ``A tissue and cell-level annotated {H\&E} and {PD-L1} histopathology
  image dataset in non-small cell lung cancer,''
\newblock {\em IEEE JBHI}, 2026.

\bibitem{ronneberger2015unet}
O.~Ronneberger et~al.,
\newblock ``{U-Net}: Convolutional networks for biomedical image
  segmentation,''
\newblock in {\em MICCAI}, 2015.

\bibitem{gamper2019pannuke}
J.~Gamper et~al.,
\newblock ``{PanNuke}: An open pan-cancer histology dataset for nuclei instance
  segmentation and classification,''
\newblock in {\em ECDP}, 2019.

\bibitem{amgad2022nucls}
M.~Amgad et~al.,
\newblock ``{NuCLS}: A scalable crowdsourcing approach and dataset for nucleus
  classification and segmentation in breast cancer,''
\newblock {\em GigaScience}, vol. 11, 2022.

\bibitem{kumar2017dataset}
N.~Kumar et~al.,
\newblock ``A dataset and a technique for generalized nuclear segmentation for
  computational pathology,''
\newblock {\em IEEE Trans. Med. Imaging}, vol. 36, no. 7, pp. 1550--1560, 2017.

\bibitem{ghahremani2022deepliif}
P.~Ghahremani et~al.,
\newblock ``Deep learning-inferred multiplex immunofluorescence for
  immunohistochemical image quantification,''
\newblock {\em Nat. Mach. Intell.}, vol. 4, pp. 401--412, 2022.

\bibitem{naumov2022endonuke}
A.~Naumov et~al.,
\newblock ``{EndoNuke}: Nuclei detection dataset for estrogen and progesterone
  stained {IHC} endometrium scans,''
\newblock {\em Data}, vol. 7, 2022,
\newblock Art. no. 75.

\bibitem{kirillov2019panoptic}
A.~Kirillov et~al.,
\newblock ``Panoptic segmentation,''
\newblock in {\em CVPR}, 2019.

\bibitem{hossain2026sinus}
M.~S. Hossain, M.~S. Rahman, and M.~Ahmed,
\newblock ``{SiNuS}: A comprehensive dataset for singular nuclei segmentation
  for {HER2} grading of breast cancer,''
\newblock {\em Data in Brief}, vol. 67, 2026,
\newblock Art. no. 112934.

\end{thebibliography}

\end{document}